\documentclass[letterpaper, 10 pt, conference]{ieeeconf}  

\IEEEoverridecommandlockouts                              

\usepackage{graphics} 
\usepackage{amsmath} 
\usepackage{amssymb}  
\usepackage{hyperref} 
\usepackage{cleveref}
\usepackage{cite}
\usepackage{graphicx}
\usepackage{booktabs}
\usepackage{multirow} 
\usepackage[table,xcdraw]{xcolor}
\usepackage{amssymb}
\usepackage{pifont}
\usepackage{caption}
\usepackage{multirow}
\usepackage{multicol}
\usepackage{booktabs}
\usepackage{hyperref}
\usepackage{caption}
\usepackage{cuted}
\DeclareCaptionLabelFormat{custom_tab}{Tab.~\thetable}
\renewcommand{\thetable}{\arabic{table}}

\newcommand\freefootnote[1]{%
  \let\thefootnote\relax%
  \footnotetext{#1}%
  \let\thefootnote\svthefootnote%
}

\title{\LARGE \bf
MoE-DP: An MoE-Enhanced Diffusion Policy for Robust Long-Horizon Robotic Manipulation with Skill Decomposition and Failure Recovery
}

\author{
Baiye Cheng$^{*\,1,4}$,
Tianhai Liang$^{*\,1}$,
Suning Huang$^{2}$,
Maanping Shao$^{1}$,\\
Feihong Zhang$^{1}$,
Botian Xu$^{1}$,
Zhengrong Xue$^{1,3}$,
Huazhe Xu$^{\dagger \,1,3}$
}

\begin{document}

\maketitle
\thispagestyle{empty}
\pagestyle{empty}

\freefootnote{${}^*\,$Equal Contribution. ${}^\dagger \,$Corresponding Author.}
\freefootnote{$^1\,$Tsinghua University. $^2\,$Stanford University. $^3\,$Shanghai Qi Zhi Institute. $^4$\,Huazhong University of Science and Technology.}


\begin{strip}
  \centering
    \includegraphics[width=0.8\textwidth]{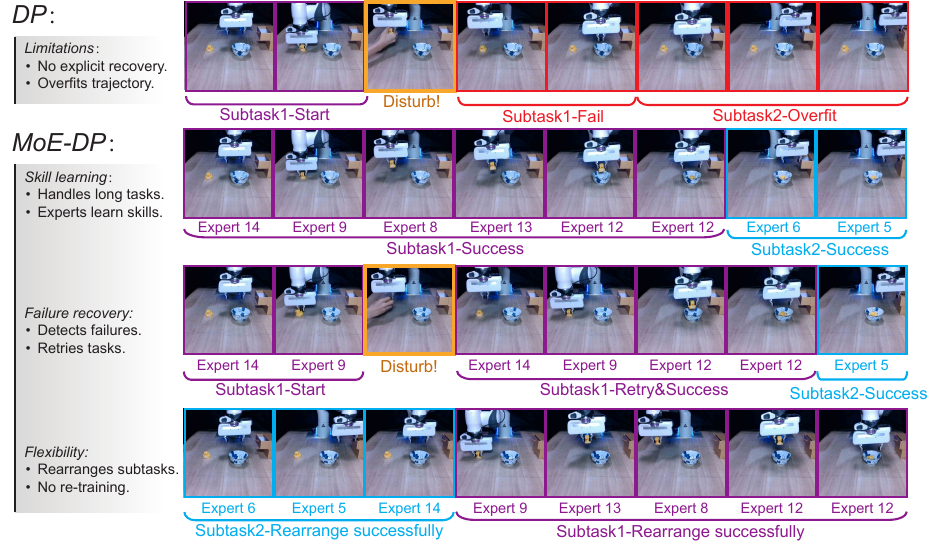}
    \captionof{figure}{\textbf{MoE-DP enables robust recovery, interpretable skill decomposition, and high-level control for long-horizon manipulation.} 
    The baseline DP fails under disturbances such as object displacement. It lacks stage awareness, cannot recover from errors, overfits to subsequent trajectories, and cascades into further failures. MoE-DP learns an interpretable skill decomposition, with experts specializing in different skills, such as approaching, grasping, and placing. MoE-DP can detect failures and reactivate the correct expert to retry failed subtasks. Task order can be flexibly rearranged by controlling the sequence of expert activations without re-training, such as executing subtask 2 before subtask 1. Colored overlays indicate expert activations and the stage of subtasks.}
    \label{fig:qualitative_results}
\end{strip}

\begin{abstract}
Diffusion policies have emerged as a powerful framework for robotic visuomotor control, yet they often lack the robustness to recover from subtask failures in long-horizon, multi-stage tasks and their learned representations of observations are often difficult to interpret. In this work, we propose the Mixture of Experts-Enhanced Diffusion Policy (MoE-DP), where the core idea is to insert a Mixture of Experts (MoE) layer between the visual encoder and the diffusion model. This layer decomposes the policy's knowledge into a set of specialized experts, which are dynamically activated to handle different phases of a task. We demonstrate through extensive experiments that MoE-DP exhibits a strong capability to recover from disturbances, significantly outperforming standard baselines in robustness. On a suite of 6 long-horizon simulation tasks, this leads to a 36\% average relative improvement in success rate under disturbed conditions. This enhanced robustness is further validated in the real world, where MoE-DP also shows significant performance gains. We further show that MoE-DP learns an interpretable skill decomposition, where distinct experts correspond to semantic task primitives (e.g., approaching, grasping). This learned structure can be leveraged for inference-time control, allowing for the rearrangement of subtasks without any re-training.
Our video and code are available at the https://moe-dp-website.github.io/MoE-DP-Website/.
\end{abstract}


\section{Introduction}

Learning visuomotor policies for robotic manipulation has become a prevailing paradigm, with recent approaches increasingly leveraging powerful generative models, such as diffusion frameworks~\cite{chi2023diffusionpolicy,black2024pi0visionlanguageactionflowmodel,lu2025h3dptriplyhierarchicaldiffusionpolicy,ze20243ddiffusionpolicygeneralizable}, to map high-dimensional observations to control actions. While these methods have demonstrated significant success in short-term tasks, they reveal a critical limitation in long-term, multi-stage scenarios: a lack of stage-awareness. When an intermediate subtask fails—for instance, an unsuccessful grasp—the policy often proceeds with the subsequent action sequence as if the failure never occurred, leading to cascading task failure. This brittleness stems from the underlying structure of the policy's learned representation, which is typically a high-dimensional and highly entangled `black box.' This opacity not only prevents the policy from recovering from local errors but also fundamentally hinders our ability to interpret how observational features map to actions, making it difficult to analyze, debug, or extend the learned behavior.

A promising architectural paradigm to address these challenges is the Mixture of Experts (MoE) framework~\cite{jacobs1991adaptive,shazeer2017outrageously,Cai_2025}, which has seen widespread success in scaling large language models~\cite{dai2024deepseekmoeultimateexpertspecialization,fedus2022switchtransformersscalingtrillion,qwen2025qwen25technicalreport,zoph2022stmoedesigningstabletransferable,qu2024llamamoev2exploringsparsity} and is now gaining traction within robotics~\cite{huang2025mentormixtureofexpertsnetworktaskoriented,chen2025gradnavvisionlanguagemodelenabled,huang2025moelocomixtureexpertsmultitask,reuss2024efficientdiffusiontransformerpolicies,li2025cloneclosedloopwholebodyhumanoid}. The core principle of MoE is to decompose a complex problem into a set of simpler ones by routing inputs to specialized sub-networks, or `experts.' This principle of modular decomposition offers a direct remedy to the limitations of conventional policies. By leveraging MoE, a single, entangled control policy can be broken down into a collection of distinct, interpretable skills, providing a clear path toward enhancing both robustness and interpretability.

In this work, we introduce the Mixture of Experts-Enhanced Diffusion Policy (MoE-DP), which integrates an MoE layer between the visual encoder and the diffusion model. MoE-DP decomposes long-horizon tasks into a set of specialized skills in an end-to-end manner. This approach enhances robustness when facing disturbance and improves interpretability by creating a clear mapping between experts and skills. Our experiments confirm that MoE-DP not only improves success rates under disturbance but also enables flexible inference-time control over learned skills.

Our contributions are threefold:
\begin{itemize}
    \item We propose MoE-DP, a novel method that demonstrates strong robustness to environmental disturbances. Evaluated across a suite of 6 long-horizon simulation tasks, MoE-DP achieves a 36\% average relative improvement in success rate under disturbed conditions where baseline methods typically fail.
    \item Through visualization, we demonstrate that MoE-DP achieves a meaningful and interpretable skill decomposition, where distinct experts are consistently activated for different semantic phases of a task (e.g., approaching, grasping, placing).
    \item We demonstrate that this learned decomposition enables inference-time control over the policy's behavior, allowing us to rearranged subtasks and generalize to new task structures without any re-training.
\end{itemize}
\section{Related Work}

\subsection{Visual Imitation Learning}
Numerous policy learning algorithms have been proposed for robotic manipulation~\cite{chi2023diffusionpolicy,zhao2023learningfinegrainedbimanualmanipulation,huang2025mentormixtureofexpertsnetworktaskoriented,huang2025dittogymlearningcontrolsoft,huang2025particleformer3dpointcloud,yuan2025hermeshumantorobotembodiedlearning}. Generative approaches like Diffusion Policies~\cite{chi2023diffusionpolicy,black2024pi0visionlanguageactionflowmodel,janner2022diffuser,sohldickstein2015deepunsupervisedlearningusing} have proven particularly effective at modeling the complex, multi-modal action distributions required for these tasks. However, their learned representations are highly entangled and lack stage-awareness, which makes the policies prone to cascading errors after intermediate failures and obscures the mapping from observation to action. In this work, we directly addresses this gap by introducing an MoE structure, which explicitly disentangles the latent space into specialized modules, thereby enhancing both the policy's robustness and the interpretability of its learned skills.

\subsection{Skill Decomposition with Mixture of Experts}
The MoE architecture has proven highly effective in large language models~\cite{dai2024deepseekmoeultimateexpertspecialization,fedus2022switchtransformersscalingtrillion,qwen2025qwen25technicalreport,zoph2022stmoedesigningstabletransferable,qu2024llamamoev2exploringsparsity,wang2024auxiliarylossfreeloadbalancingstrategy} and multi-task robotics~\cite{huang2025mentormixtureofexpertsnetworktaskoriented,huang2025moelocomixtureexpertsmultitask,chen2025gradnavvisionlanguagemodelenabled,reuss2024efficientdiffusiontransformerpolicies,li2025cloneclosedloopwholebodyhumanoid}. Building on these successes, our work adapts this paradigm not for inter-task assignment, but for fine-grained \emph{skill decomposition} within a single long-horizon task. Unlike traditional two-stage methods that discover skills separately before policy integration~\cite{zhu2022bottomupskilldiscoveryunsegmented,xu2023xskillcrossembodimentskill,liang2024skilldiffuserinterpretablehierarchicalplanning}, we integrate an MoE layer directly into the policy for end-to-end learning. This yields a set of interpretable, task-optimized skills that can be flexibly controlled at inference time to alter the execution order of subtasks.

\begin{figure*}[t!]
    \centering
    \vspace{3mm}
    \includegraphics[width=0.8\textwidth]{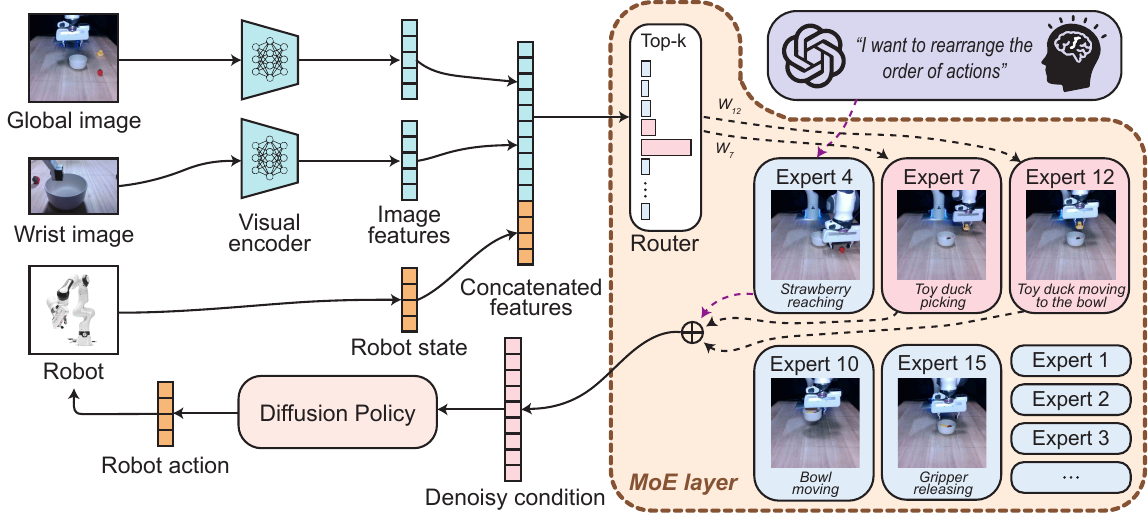}
    \caption{\textbf{Overview of MoE-DP with high-level guidance.} In its \textbf{autonomous mode}, the system encodes observation inputs (images and robot state) into a feature vector, which is then fed to an MoE layer. The MoE's router automatically selects the appropriate expert for the current observation. The output of the selected expert then serves as a conditioning input for the Diffusion Policy during action generation. While the router typically operates autonomously, the architecture supports \textbf{high-level control}: an external agent, such as a human operator or a Vision-Language Model (VLM), can guide the policy by \textbf{overriding the router's default selection}. This capability enables flexible behaviors, such as reordering subtasks to generalize to novel sequences not seen during training.}
    \label{fig:overview}
\end{figure*}


\subsection{Inference-Time Control of Policy Behavior}

Modifying a policy's behavior at inference time without re-training is a key challenge in robotics~\cite{uehara2025inferencetimealignmentdiffusionmodels,wang2025inferencetimepolicysteeringhuman,ajay2023conditionalgenerativemodelingneed,du2021modelbasedplanningenergy,gkanatsios2024energybasedmodelszeroshotplanners,reuss2023goalconditionedimitationlearningusing}. Common strategies rely on external modules, such as safety value functions~\cite{liu2024multitaskinteractiverobotfleet,nakamura2025generalizingsafetycollisionavoidancelatentspace} or gradient-based guidance from dynamics models~\cite{du2025dynaguidesteeringdiffusionpolices}, which typically require separate training. In contrast, our work provides an \textbf{intrinsic control mechanism} by directly leveraging the policy's learned MoE structure. The resulting skill decomposition offers interpretable ``handles" for high-level control, as each expert corresponds to a clear semantic skill, enabling dynamic control over the execution flow without requiring auxiliary models.
\section{Method}

To learn a more structured and human-interpretable representation, we introduce an MoE layer to dynamically process the latent features obtained from the observation encoder. This strategic integration transforms the policy's conditioning from a static feature vector into a dynamic, structured representation. This learned decomposition is the key to enabling two critical capabilities: robust recovery from intermediate task failures and flexible, high-level control over the policy's behavior at inference time.

\subsection{Preliminaries}

\subsubsection{Visuomotor Diffusion Policy}
A diffusion policy models the visuomotor mapping $p(A_t|O_t)$ as a conditional denoising process~\cite{ho2020denoisingdiffusionprobabilisticmodels,welling2011bayesian,rissanen2022generative}. An encoder first processes a history of observations into a latent vector $\mathbf{z}_t = \text{Encoder}(O_{t-T_o+1:t})$. The policy then generates an action sequence by iteratively refining Gaussian noise over multiple steps using a learned denoising network $\epsilon_{\theta}(A^k_t, \mathbf{z}_t, k)$. The training objective is a mean-squared error loss to predict the noise added to ground-truth actions:
\begin{equation} \label{eq:loss_diff}
\mathcal{L}_{\text{diff}} = \mathbb{E}_{A_{t:t+T_a}, \epsilon, k} \left[ \left\| \epsilon - \epsilon_{\theta}(A^0_t + \sigma_k \epsilon \mid \mathbf{z}_t, k) \right\|^2 \right]
\end{equation}

\subsubsection{Mixture of Experts}
The Mixture of Experts (MoE) architecture enables conditional computation through a set of $N$ ``expert" networks (typically feed-forward networks) and a ``router". For an input $\mathbf{x}$, the router computes a weighting distribution, $g(\mathbf{x})$, which is typically generated by applying a softmax function to the output of a linear layer. The final output is a weighted combination of the outputs from the activated experts (often only the Top-k):
\begin{equation} \label{eq:moe_output_prelim}
\text{MoE}(\mathbf{x}) = \sum_{i \in \text{Top-k}(\mathbf{x})} g(\mathbf{x})_i \cdot E_i(\mathbf{x})
\end{equation}
This allows for a significant increase in model capacity without a proportional rise in computational cost, as individual experts learn specialized functions.

\subsection{MoE-Enhanced Diffusion Policy (MoE-DP)}

Our core contribution is the MoE-DP, which integrates an MoE~\cite{jacobs1991adaptive,shazeer2017outrageously} layer into the Diffusion Policy~\cite{chi2023diffusionpolicy} framework. The implementation is built upon the original Diffusion Policy architecture; it takes as input a global camera view, a wrist-mounted camera view, and the robot's proprioceptive state, processed by a standard ResNet18~\cite{he2015deepresiduallearningimage} visual encoder that is trained from scratch. By interposing the MoE layer between this visual encoder and the diffusion model, MoE-DP encourages the policy to learn a decomposed skill representation. This architecture achieves two primary goals: enhancing the policy’s robustness in multi-stage tasks by making it stage-aware, and improving the interpretability of its internal representations by encouraging specialization.

\subsubsection{MoE Architecture}

Our primary architectural modification is to insert an MoE layer to process the latent feature vector $\mathbf{z}_t$ from the visual encoder. Instead of directly conditioning the diffusion model on latent feature, the MoE layer dynamically routes $\mathbf{z}_t$ through a set of specialized expert networks to produce a refined conditioning feature. Each expert, $\{E_i\}_{i=1}^N$, is a multi-layer perceptron (MLP) designed to specialize in a specific phase of the manipulation task. A lightweight gating network, the \textbf{router}, computes routing weights for these $N$ experts:
\begin{equation} \label{eq:gating}
g_t = \text{Softmax}(W_g \mathbf{z}_t)
\end{equation}
where $W_g$ is a learned weight matrix. Instead of a dense combination, we employ a sparse, Top-k routing strategy. The final conditioning vector $\mathbf{z}'_t$ is a weighted combination of the outputs of only the top-$k$ selected experts:
\begin{equation} \label{eq:moe_output}
\mathbf{z}'_t = \sum_{i \in \text{Top-k}(g_t)} g_{t,i} \cdot E_i(\mathbf{z}_t)
\end{equation}
This MoE-enhanced feature $\mathbf{z}'_t$ then serves as the condition for the diffusion model $\epsilon_{\theta}$. This design encourages different experts to specialize in distinct phases of a task (skill decomposition). The router's gating mechanism can thus re-activate an appropriate expert to retry a failed subtask, promoting stage-aware robustness.

\subsubsection{Training Objective}

A key challenge when training MoE models is to prevent \textit{router collapse} \cite{lepikhin2020gshardscalinggiantmodels , fedus2022switchtransformersscalingtrillion}, where the gating network learns to always select only a few dominant experts, leaving others untrained. To ensure all experts are utilized and learn meaningful specializations, we introduce an auxiliary loss, $\mathcal{L}_{\text{aux}}$, which is critical for achieving balanced and specialized expert usage.

This auxiliary loss consists of two terms. The first is a \textbf{load-balancing loss}\cite{lepikhin2020gshardscalinggiantmodels , fedus2022switchtransformersscalingtrillion}, which encourages a more uniform load distribution across experts, preventing the router from consistently selecting only a few experts while leaving others untrained. This loss is computed as the scaled dot-product between the fraction of tokens dispatched to each expert and the fraction of router probability for them:
\begin{equation} \label{eq:loss_load}
\mathcal{L}_{\text{load}} = N \sum_{i=1}^{N} f_i \cdot P_i
\end{equation}
where $N$ is the total number of experts. Here, $f_i$ is the fraction of samples in the batch $\mathcal{B}$ dispatched to expert $i$:
\begin{equation}
f_i = \frac{1}{B} \sum_{t \in \mathcal{B}} \mathbf{1}\{\text{argmax}_j(p_{t,j}) = i\}
\end{equation}
and $P_i$ is the average router probability for expert $i$ across the batch:
\begin{equation}
P_i = \frac{1}{B} \sum_{t \in \mathcal{B}} p_{t,i}
\end{equation}
where $B$ is the batch size and $p_{t,i}$ is the router's softmax output probability for sample $t$ to expert $i$.

The second term is an \textbf{entropy loss}, which promotes a more structured division of labor among the experts and enhances their specialization. By encouraging the router to make confident, low-entropy (i.e., high-probability) assignments for each individual sample, it sharpens the role of each expert. This specialization is crucial for downstream goals: it improves interpretability by creating a clearer mapping between experts and skills, and it enables \textbf{high-level behavioral control}. The loss is the mean entropy of the router's output distribution over the batch:
\begin{equation} \label{eq:loss_entropy}
\mathcal{L}_{\text{entropy}} = - \frac{1}{B} \sum_{t=1}^{B} \sum_{i=1}^{N} p_{t,i} \log(p_{t,i} + \epsilon)
\end{equation}
where $\epsilon$ is a small constant for numerical stability.

The training of MoE-DP model is guided by a composite objective function. This function combines the standard diffusion loss with auxiliary terms that encourage balanced and specialized expert utilization. The complete training objective is a weighted sum of these components:
\begin{equation} \label{eq:loss_total}
\mathcal{L} = \mathcal{L}_{\text{diff}} + \underbrace{\lambda \mathcal{L}_{\text{load}} + \beta \mathcal{L}_{\text{entropy}}}_{\text{Auxiliary Loss}}
\end{equation}
where $\mathcal{L}_{\text{diff}}$ is the standard diffusion denoising loss. The \textbf{auxiliary loss} term consists of the load-balancing loss ($\mathcal{L}_{\text{load}}$) and the entropy loss ($\mathcal{L}_{\text{entropy}}$), weighted by their respective coefficients $\lambda$ and $\beta$. This combined objective guides the model to learn a set of specialized experts for different subtasks while accurately modeling the action distribution.

\subsection{Skill Decomposition and Inference-Time Control of Behavior}

Leveraging the auxiliary loss, MoE-DP model learns a disentangled and interpretable representation of skills, enabling effective decomposition of long-horizon, multi-stage tasks. This is achieved through the synergy of its two components. The load-balancing loss encourages a more uniform load distribution across experts, ensuring a wider range of experts are utilized during training. Meanwhile, the entropy loss incentivizes the router to produce low-entropy, high-confidence probability distributions. This ensures that for any given observation, the router’s output is sharply peaked, meaning one expert is typically selected with a probability approaching 1, while all others receive near-zero probability.
\begin{figure*}[t]
    \centering
    \vspace{3mm}
    \includegraphics[width=0.8\textwidth]{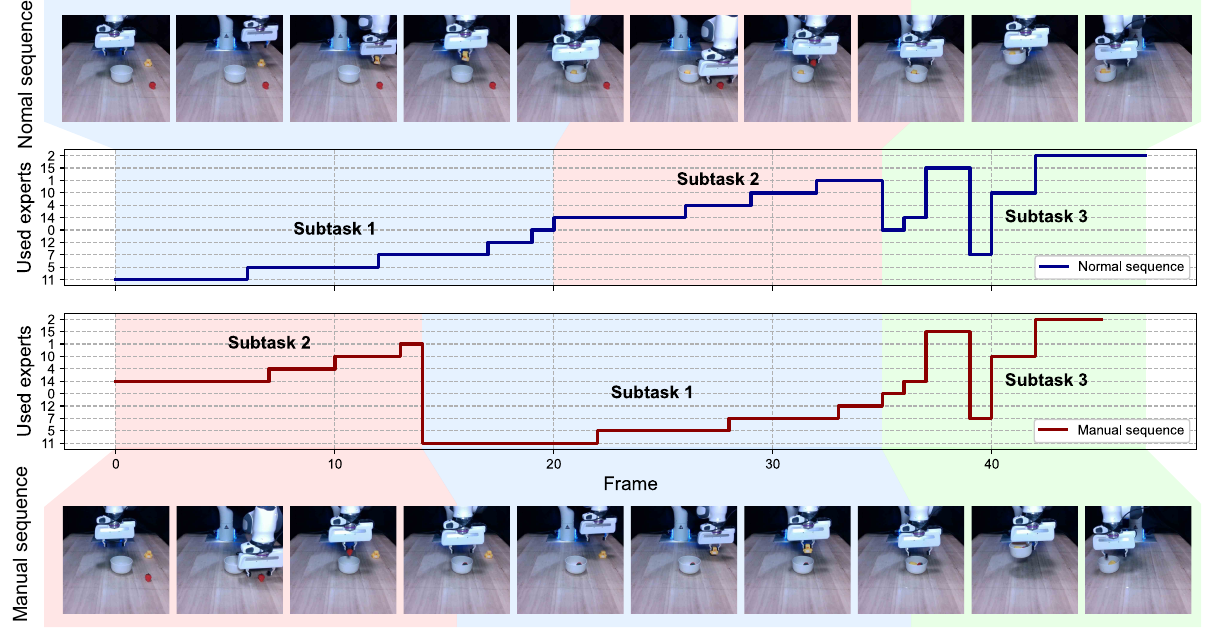}
    \caption{\textbf{Inference-time control via compositional skill decomposition.} We demonstrate that MoE-DP learns modular and reusable skills that can be flexibly recombined to form novel behaviors without re-training. \textbf{(Top)} When executing a task in its demonstrated order, the policy decomposes the process into three distinct subtasks—Subtask 1 (picking the yellow duck), Subtask 2 (picking the strawberry), and Subtask 3 (moving the bowl)—each invoking a consistent sequence of expert activations. \textbf{(Bottom)} At inference time, we manually command a novel sequence by altering the subtask order (Subtask 2 followed by Subtask 1). The policy successfully executes this new task by reordering the learned skill modules. Crucially, the expert activation pattern for an individual subtask (e.g., Subtask 1) remains consistent across both scenarios, proving that MoE-DP learns truly compositional skills that enable generalization to new task structures.}
    \label{fig:compositional_skills}
\end{figure*}

And this emergent property is the foundation for two powerful capabilities of MoE-DP. First, it enables a clear and interpretable \textbf{skill decomposition}. Because a single, dominant expert is active during each phase of a manipulation task, a direct correspondence emerges between specific experts and distinct, semantic skills (e.g., approaching, grasping, placing). This transparency allows us to analyze and understand the policy's decision-making process in a way that is impossible with conventional architectures.
\begin{table}[]
\vspace{2mm}
\caption{\textbf{Simulation task results.} We compare the performance of MoE-DP against the baseline across all simulation tasks under both nominal (\checkmark) and disturbed (\ding{55}) execution conditions. The Disturbance (Object) column indicates the action of resetting the object to its initial position after it has been successfully grasped. Bolded numbers indicate the highest success rate for each condition.}
\label{tab:sim_results}
\renewcommand{\arraystretch}{1.25}
\begin{tabular}{c|c|cc}
\hline
\multirow{2}{*}{Task}                  & \multicolumn{1}{c|}{\multirow{2}{*}{Disturbance (Object)}} & \multicolumn{2}{c}{Method}        \\ \cline{3-4} 
                                       & \multicolumn{1}{c|}{}                                      & DP             & \textbf{MoE-DP} \\ \hline
\multirow{2}{*}{Hammer Cleanup T0}     & \checkmark                                                 & \textbf{86.7}  & 84.0             \\
                                       & \ding{55}(Hammer)                                          & 17.3           & \textbf{33.3}    \\ \hline
\multirow{3}{*}{Kitchen T0}            & \checkmark                                                 & \textbf{93.3}  & 80.7             \\
                                       & \ding{55}(Cube)                                            & 32.0           & \textbf{48.7}    \\
                                       & \ding{55}(Pot)                                             & 18.7           & \textbf{62.7}    \\ \hline
\multirow{2}{*}{Coffee Preparation T0} & \checkmark                                                 & \textbf{52.7}  & 51.3             \\
                                       & \ding{55}(Mug)                                             & 10.0           & \textbf{16.0}    \\ \hline
\multirow{2}{*}{Mug Cleanup T0}        & \checkmark                                                 & 59.3           & \textbf{62.0}    \\
                                       & \ding{55}(Mug)                                             & 32.7           & \textbf{35.3}    \\ \hline
\multirow{2}{*}{Kitchen Cleanup T0}    & \checkmark                                                 & 52.7           & \textbf{96.7}   \\
                                       & \ding{55}(Hammer)                                          & 12.0           & \textbf{82.7}    \\ \hline
\multirow{2}{*}{Table Cleanup T0}      & \checkmark                                                 & \textbf{100.0} & \textbf{100.0}   \\
                                       & \ding{55}(Cube)                                            & 2.0            & \textbf{98.0}    \\ \hline
\multirow{2}{*}{\textbf{Average}}      & \checkmark                                                 & 74.1           & \textbf{79.1}    \\
                                       & \ding{55}                                                    & 17.8           & \textbf{53.8}    \\ \hline
\end{tabular}
\end{table}

Second, this explicit expert-skill mapping enables direct \textbf{inference-time control of behavior}. Since the choice of expert dictates the subsequent action sequence, we can externally manipulate the router's output to command specific behaviors. This opens up possibilities for hierarchical control, where a high-level agent—such as a human operator or a VLM—can direct the policy's execution flow. A compelling application of this is generalizing beyond the training data's fixed task sequences. For example, consider a policy trained exclusively on demonstrations that execute subtask A before subtask B. While a standard policy would be confined to this order, MoE-DP allows a VLM, given the instruction ``Do B then A," to simply alter the activation order of the corresponding experts. This allows the policy to generalize to novel task structures not seen in the training data. And we will talk this capability in our experiments part~\ref{sec:skill_decomp}

\section{Experiments}

\begin{table}[]
\vspace{2mm}
\caption{\textbf{Real-world task results.} We compare the performance of MoE-DP against the baseline across all real-world tasks under both nominal (\checkmark) and disturbed (\ding{55}) execution conditions. The Disturbance (Object) column indicates the action of resetting the object to its initial position after it has been successfully grasped. Bolded numbers indicate the highest success rate for
each condition.}
\renewcommand{\arraystretch}{1.25}
\begin{tabular}{c|c|cc}
\toprule
                      & \multicolumn{1}{c|}{}    & \multicolumn{2}{c}{Method}     \\ \cline{3-4} 
\multirow{-2}{*}{Task Name}                        & \multicolumn{1}{c|}{\multirow{-2}{*}{Disturbance (Object)}} & DP   & \textbf{MoE\_DP} \\ \hline
                      & {\color[HTML]{333333} \checkmark} & 85.0          & \textbf{90.0}  \\
                      & \ding{55}(Green Cube)            & 5.0           & \textbf{60.0}  \\
\multirow{-3}{*}{Pick two cube}             & \ding{55}(Red Cube)                                                 & 75.0 & \textbf{90.0}    \\ \hline
                      & {\color[HTML]{333333} \checkmark} & \textbf{90.0} & \textbf{90.0}  \\
\multirow{-2}{*}{Duck place drawer close}  & \ding{55}(Cube)                                                     & 5.0  & \textbf{70.0}    \\ \hline
                      & {\color[HTML]{333333} \checkmark} & 95.0          & \textbf{100.0} \\
                      & \ding{55}(Duck)                  & 20.0          & \textbf{55.0}  \\
\multirow{-3}{*}{Duck place bowl transport} & \ding{55}(Strawberry)                                               & 35.0 & \textbf{80.0}    \\ \hline
                      & {\color[HTML]{333333} \checkmark} & 90.0          & \textbf{93.3}  \\
\multirow{-2}{*}{\textbf{Average}} & \ding{55}                        & 28.0          & \textbf{71.0}  \\ \bottomrule
\end{tabular}
\label{tab:real_results}
\end{table}

We evaluate MoE-DP on a series of long-horizon manipulation tasks in both simulation and the real world. We use Diffusion Policy (DP) as our baseline. To ensure a fair comparison, the network capacity of MoE-DP does not differ significantly from the baseline. The experiments are designed to investigate three central questions:

\begin{itemize}
    \item Does MoE-DP outperform a standard diffusion policy baseline in long-horizon, multi-stage tasks, especially under disturbed conditions? (Sec.~\ref{sec:sim_exp}, \ref{sec:real_exp})
    \item Does the MoE-DP learn a meaningful skill decomposition, where distinct experts consistently activate for different subtasks or behaviors? (Sec.~\ref{sec:skill_decomp})
    \item Can the learned skill decomposition be leveraged to flexibly control the policy's behavior at inference time, for instance, to rearranged subtasks? (Sec.~\ref{sec:control})
    \item How does the auxiliary loss contribute to the model's performance and skill decomposition? (Sec.~\ref{sec:ablation})
    
\end{itemize}

\subsection{Simulation Experiments}
\label{sec:sim_exp}

\noindent\textbf{Experimental Setup.}
Inspired by the MimicGen~\cite{mandlekar2023mimicgendatagenerationscalable} benchmark, we constructed a suite of 6 long-horizon tasks in a simulation environment. These tasks involve multi-stage, multi-object manipulation and are specifically designed to challenge a policy's stage-awareness and its ability to recover from subtask failures. We evaluate all policies under two conditions: (1) \textit{nominal execution}, where tasks proceed without disturbance, and (2) \textit{disturbed execution}, a condition designed to explicitly test the policy's robustness and recovery capabilities. In this setup, we programmatically induce a failure within a specific subtask. For instance, after the robot arm successfully grasps an object but before it can place it at the target location, we reset the object back to its initial position. This forces the policy to recognize the state discrepancy and re-attempt the subtask. For each simulation task, we collect 100 expert demonstrations. To accelerate training, we use a batch size of 128, while all other hyperparameters
and training settings are kept consistent with the original Diffusion Policy implementation. We run 3 seeds for each experiment. For each seed, we train policies for 500 epochs, evaluating with 50 rollouts every 10 epochs. We report the maximum success rate achieved during training, averaged across the three seeds.

\begin{table}[]
\caption{\textbf{MoE hyperparameter settings.} Due to the unique characteristics of each task and its training data, optimal hyperparameter settings were found empirically. `expert\_num' denotes the total number of experts in the MoE architecture; `top\_k' specifies the number of experts activated by the router at each forward pass; $\lambda$ is the weight for the load-balancing loss, and $\beta$ is the weight for the entropy loss.}
\begin{tabular}{@{}ccccc@{}}
\toprule
\multicolumn{1}{c|}{Task Name}                     & $\#$experts & top-k & $\lambda$ & $\beta$ \\ \midrule
\multicolumn{5}{c}{Simulation}                                                              \\ \midrule
\multicolumn{1}{c|}{Hammer Cleanup T0}      & 8           & 2      & 0.1       & 0.01     \\
\multicolumn{1}{c|}{Kitchen T0}              & 16          & 2      & 0.1       & 0.007    \\
\multicolumn{1}{c|}{Coffee Preparation T0}          & 16          & 4      & 0.1       & 0.01     \\
\multicolumn{1}{c|}{Mug Cleanup T0}        & 16          & 4      & 0.1       & 0.01     \\
\multicolumn{1}{c|}{Kitchen Cleanup T0}        & 16          & 2      & 0.1       & 0.01     \\
\multicolumn{1}{c|}{Table Cleanup T0}         & 16          & 2      & 0.1       & 0.01     \\ \midrule
\multicolumn{5}{c}{Real-world}                                                              \\ \midrule
\multicolumn{1}{c|}{Pick two cube}          & 16          & 2      & 0.1       & 0.03     \\
\multicolumn{1}{c|}{Duck place drawer close} & 16          & 2      & 0.1       & 0.04     \\
\multicolumn{1}{c|}{Duck place bowl transport} & 16          & 2      & 0.1       & 0.03     \\ \bottomrule
\end{tabular}
\label{tab:moe_hyperparameters}
\end{table}

\noindent\textbf{Results.}
As shown in Tab.~\ref{tab:sim_results}, under \textit{nominal execution} conditions, the performance of MoE-DP is comparable to the baseline. However, the advantage of MoE-DP becomes evident in the \textit{disturbed execution} scenario, where MoE-DP's success rate is 36\% higher than the baseline's. This result demonstrates MoE-DP's robustness in the face of disturbances. During the tuning process, we observed that due to the distinct characteristics of the data for each task, different tasks require specific MoE hyperparameter configurations to achieve optimal performance, as detailed in Tab.~\ref{tab:moe_hyperparameters}.

\begin{table}[]
\vspace{2mm}
\caption{\textbf{Generalization to novel task sequences via inference-time control.} We quantitatively evaluate the policy's ability to execute subtasks in novel orders not seen during training. The execution flow was guided at inference time by either a human operator (\texttt{Human\_Control}) or a VLM (\texttt{VLM\_Control}). Success rates are reported over 10 trials for each task, demonstrating the flexibility of the learned compositional skills.}
\begin{tabular*}{\columnwidth}{@{\extracolsep{\fill}}c|cc@{}}
\toprule
Task Name & Human\_Control & VLM\_Control \\ \midrule
Pick two cube & 7/10 & 6/10 \\
Duck place drawer close & 9/10 & 5/10 \\
Duck place bowl transport & 9/10 & 5/10 \\ \bottomrule
\end{tabular*}
\label{tab:inference_control_results}
\end{table}

\subsection{Real-World Experiments}
\label{sec:real_exp}

\noindent\textbf{Experimental Setup.}
We deploy MoE-DP on a Franka Emika robot, using two Intel RealSense D435i cameras (one stationary for a global view and one wrist-mounted) for visual feedback. We designed three long-horizon manipulation tasks to evaluate performance on physical hardware:
\begin{itemize}
    \item \textbf{Duck place drawer close:} Place a rubber duck into a bowl and subsequently close a nearby drawer.
    \item \textbf{Pick two cube:} Pick and place two colored blocks into a designated container in a predefined order.
    \item \textbf{Duck place bowl transport:} Place both a duck and a strawberry into a bowl, and then transport the bowl to a target location.
\end{itemize}
These tasks require the policy to chain together fine-grained manipulation skills and reason about the temporal progression of the task.


\begin{figure}[t]
    \centering
    \includegraphics[width=0.48\textwidth]{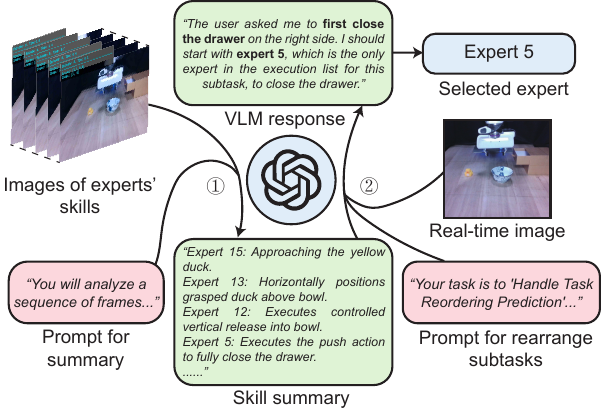}
\caption{\textbf{Overview of the VLM-based planning and control framework.} Our system leverages a VLM for high-level task planning in two stages. First, at the \textit{skill summarization} (\ding{172}) stage, the VLM builds a textual knowledge base of the robot's capabilities by analyzing annotated frames from a demonstration that follows the same execution sequence as the training data. Second, at the \textit{task execution} stage (\ding{173}), the VLM uses this knowledge, a high-level goal, and a real-time image to reason about the current task stage and predict the appropriate expert to activate. This hierarchical architecture enables the system to dynamically plan and rearrange the order of subtasks without any re-training, translating abstract goals into concrete robotic actions.}
\label{fig:vlm_framework}
\end{figure}

\noindent\textbf{Results.}
We evaluate the trained models over 20 test trials for each task. As shown in Tab.~\ref{tab:real_results}, the real-world results directly mirrored the simulation findings. While both policies performed comparably under nominal conditions, MoE-DP again demonstrated significantly higher robustness when subjected to manual disturbance, reliably recovering from induced failures where the baseline could not. This improved real-world reliability is a direct result of the structured behavior promoted by the MoE architecture, leading to more predictable and interpretable action sequences as different experts specialize in distinct task stages.

\subsection{Analysis of Skill Decomposition}
\label{sec:skill_decomp}

To verify that MoE-DP learns a meaningful skill decomposition, we visualize expert activations during the inference phase of our tasks. As visualized in the top panel of Fig.~\ref{fig:compositional_skills}, MoE-DP, guided by the auxiliary loss, successfully decomposes long-horizon, multi-stage tasks. We observe a clear and consistent pattern where different experts are systematically activated for distinct execution stages. For example, during the subtask of placing a duck into a bowl, a clear division of labor emerges: one expert becomes primarily active for `approaching the duck', a second expert handles the `grasping' action, and a third takes over for the subsequent phase of `moving it into the bowl'. This fine-grained specialization confirms that MoE-DP disentangles complex behaviors into interpretable, semantically consistent primitives, showcasing a robust and meaningful skill decomposition.

\subsection{Inference-Time Control of Behavior}
\label{sec:control}

Additionally, we investigate whether the learned skill decomposition enables explicit, high-level control over the policy's execution flow. As qualitatively demonstrated in Fig.~\ref{fig:compositional_skills}, MoE-DP allows for the flexible recomposition of learned skills. This control can be exerted at inference time by an external agent, such as a human operator or a VLM. We propose a hierarchical framework for VLM-based control, detailed in Fig.~\ref{fig:vlm_framework}, where the VLM reasons about the task and directs the policy by selecting the appropriate expert sequence. This allows the robot to execute subtasks in novel orders (e.g., B $\rightarrow$ A $\rightarrow$ C) not seen in the training data, without any re-training. We quantitatively evaluate this capability in Tab.~\ref{tab:inference_control_results}, reporting success rates over 10 trials for both human and VLM control. The results confirm the flexibility afforded by the compositional skills, though we note that the VLM's performance is lower than that of direct human control due to its limitations in spatial understanding and precise progress recognition. Ultimately, this demonstrates that MoE-DP not only decomposes tasks into reusable skills but also allows for their flexible recomposition, enabling generalization to new task structures.

\subsection{Ablation Study on Auxiliary Loss}
\label{sec:ablation}


\begin{table}[]
\vspace{2mm}
\caption{\textbf{Ablation study on the auxiliary loss.} Bold numbers denote the best result in each row, and underlined numbers denote the second best. Across all tasks, the combination (LE) of the load-balance loss (L) and entropy loss (E) gives the best performance.}
\renewcommand{\arraystretch}{1.25}
\begin{tabular}{c|c|cccc}
\toprule
                                     &                          & \multicolumn{4}{c}{Method}                                                                            \\ \cline{3-6} 
\multirow{-2}{*}{Task Ablation} & \multirow{-2}{*}{\begin{tabular}[c]{@{}c@{}}Disturbance \\ (Object)\end{tabular}} & LE & L & E & N \\ \hline
                                     & {\color[HTML]{333333} \checkmark} & \textbf{84.0}  & 82.0 & \underline{83.0}          & 82.7                                  \\
\multirow{-2}{*}{Hammer Cleanup T0}     & \ding{55}(Hammer)                & \textbf{33.3}  & 19.7 & \underline{26.5}          & 22.7                                  \\ \hline

                                     & {\color[HTML]{333333} \checkmark} & \textbf{96.7} & 86.0 & \underline{93.0}          & 86.0                                  \\
\multirow{-2}{*}{Kitchen Cleanup T0}    & \ding{55}(Hammer)                & 
\textbf{82.7}  & 65.3 & \underline{74.7}          & 54.7                                  \\ \hline
                                     & {\color[HTML]{333333} \checkmark} & \textbf{100.0} & 96.7 & 98.3          & \underline{99.3}                                  \\
\multirow{-2}{*}{Table Cleanup T0}      & \ding{55}(Cube)                  & \textbf{98.0}  & 33.3 & \underline{64.7}          & 44.0                                  \\ \toprule
\end{tabular}
\label{tab:ablation_results}
\end{table}

Finally, to isolate the contributions of the two components of the auxiliary loss, we conduct an ablation study with results summarized in Tab.~\ref{tab:ablation_results}. We evaluate four variants of the auxiliary loss: a version with no auxiliary loss (\textbf{N}), a version with only the load-balancing loss (\textbf{L}), a version with only the entropy loss (\textbf{E}), and our proposed model which combines the load-balancing and entropy losses (\textbf{LE}). We compare these variants on both task performance and the quality of the learned skill decomposition.

Our findings confirm that the combination of both the load-balancing and entropy losses yields the best overall task success rate. Regarding skill decomposition, we made several key observations. When using only the load-balancing loss, we found that while all experts were utilized during training, no clear expert specialization emerged. The router's output distribution remained relatively uniform, which prevents a clear mapping from experts to specific behaviors and hinders effective inference-time control. Conversely, when using only the entropy loss, the model suffered from severe router collapse, where the gating network quickly learned to activate only a small subset of experts, leaving the majority completely untrained. The full auxiliary loss, combining both terms, successfully addresses both issues. The load-balancing term ensures all experts participate in the training process, while the entropy term encourages each expert to develop a sharp, specialized function. This synergy is crucial for learning the robust and interpretable skill decomposition that is central to MoE-DP's success.

\section{Conclusion}

In this work, we addressed the critical challenges of robustness and interpretability in diffusion policies by introducing the MoE-DP. MoE-DP learns to decompose complex behaviors into a set of specialized and interpretable skills. This novel structure significantly enhances robustness, enabling reliable recovery from subtask failures where standard policies fail. Furthermore, MoE-DP learned decomposition provides a handle for inference-time control, allowing for the flexible recombination of skills to execute novel task sequences without re-training.

\noindent\textbf{Limitations and Future Work.}
A primary limitation of MoE-DP is that, due to the varying data structures and characteristics of different tasks, identifying the optimal MoE architecture and related hyperparameters still requires empirical tuning. Future work could investigate adaptive or automated strategies for selecting MoE configurations, enabling more robust and generalizable robotic agents.




\bibliographystyle{IEEEtran} 
\clearpage
\bibliography{reference}  



\end{document}